\pdfoutput=1

\documentclass[11pt]{article}

\usepackage{acl}

\usepackage{times}
\usepackage{latexsym}

\usepackage[T1]{fontenc}

\usepackage[utf8]{inputenc}

\usepackage{microtype}

\usepackage{inconsolata}

\usepackage{enumitem}
\usepackage{booktabs}
\usepackage{graphicx}
\usepackage{amsfonts}
\usepackage{amssymb}
\usepackage{amsmath}
\usepackage{listings}

\usepackage{amssymb}
\usepackage{pifont}
\newcommand{\cmark}{\ding{51}}%
\newcommand{\xmark}{\ding{55}}%

\usepackage{xcolor}

\newcommand{\booksql}{\textbf{BookSQL}}
\newcommand{\booksqlNormal}{{BookSQL}}
\newcommand{\textsql}{{Text-to-SQL}}

\title{BookSQL: A Large Scale Text-to-SQL Dataset for Accounting Domain}

\author{Rahul Kumar\footnotemark[2] \qquad Amar Raja Dibbu\footnotemark[2] \qquad Shrutendra Harsola\footnotemark[1] \\
\textbf{Vignesh Subrahmaniam}\footnotemark[1] \qquad \textbf{Ashutosh Modi}\footnotemark[2] \\
\footnotemark[2]\ \ Indian Institute of Technology Kanpur (IIT Kanpur) \qquad \footnotemark[1]\ \ Intuit\\
\texttt{\{rahulkumar21,amard21\}@iitk.ac.in} \\ \texttt{\{shrutendra\_harsola,vignesh\_subrahmaniam\}@intuit.com} \\ 
\texttt{\{ashutoshm\}@cse.iitk.ac.in} 
}

\begin{document}
\maketitle

\begin{abstract}
Several large-scale datasets (e.g., WikiSQL, Spider) for developing natural language interfaces to databases have recently been proposed. These datasets cover a wide breadth of domains but fall short on some essential domains, such as finance and accounting. Given that accounting databases are used worldwide, particularly by non-technical people, there is an imminent need to develop models that could help extract information from accounting databases via natural language queries. In this resource paper, we aim to fill this gap by proposing a new large-scale Text-to-SQL dataset for the accounting and financial domain: BookSQL. The dataset consists of 100k  natural language queries-SQL pairs, and accounting databases of 1 million records. We experiment with and analyze existing state-of-the-art models (including GPT-4) for the Text-to-SQL task on BookSQL. We find significant performance gaps, thus pointing towards developing more focused models for this domain. 
\end{abstract}

\section{Introduction} \label{sec:intro}

Relational databases are pervasive in all modern-day organizations, from financial establishments to educational institutes. Typically, query languages such as SQL are used to extract the required data from relational databases. However, formulating queries in SQL needs mastery of the language itself; consequently, this excludes people (particularly those without technical background, e.g., financial accountants) who do not know SQL from using databases. It is imperative to develop techniques to address the research question, can relational databases be queried using natural language? In this paper, we take a step toward this goal; in particular, we explore if one could develop a natural language interface for accounting databases. In recent years, several large-scale general-purpose datasets \cite{deng-etal-2022-recent-advances-in-text-to-sql} have been proposed for developing \textsql\  systems\footnote{By \textsql\  system we refer to a system that, given a natural language query, automatically retrieves the desired information from a database or multiple databases by converting a natural language query to SQL query as an intermediate representation.}, such as Spider \cite{spider234} and WikiSQL \cite{wikiSQL}. 
\begin{table}[t]
    \centering
    \begin{tabular}{ccc}
        \toprule
        \textbf{Model} & \textbf{Spider} & \textbf{\booksql}\\     
        \midrule
        UniSAr & 70\% & 3.8\% \\
        SEDE & 63.2\% & 0.0\% \\
        RESDSQL & 80.5\% & 10.8\% \\
\bottomrule
\end{tabular}
\caption{Performance (Exact Match Accuracy (c.f. \S\ref{sec:experiments})) of pre-trained SOTA \textsql\ models on Spider and the proposed \booksql\ dataset. As can be observed existing models have very poor performance on \booksql\ indicative of poor domain generalization.} 
\label{tab:sota-models}
\vspace{-4mm}
\end{table}
Such datasets,\footnote{By \textsql\  dataset we refer to a dataset having both the natural language queries with corresponding SQL formulation and correct answers along with the corresponding database against which queries are fired} though cross-domain, are still not suitable for developing systems that could address real-world business use cases, such as accessing accounting databases via natural language interfaces. The primary reason is that these large-scale datasets have a considerable breadth regarding types of domains. However, they either lack certain domains (such as accounting) or have limited data and query types for specific domains (e.g., financial, sales, and marketing). In this paper, we try to address this gap by proposing a large-scale \textsql\ dataset (called \booksql) for the accounting and business domain. We collaborate with financial experts to create a dataset that reflects actual accounting databases used in the industry.  

\begin{table*}[t!]
\centering
\renewcommand{\arraystretch}{1.0}
\setlength\tabcolsep{5pt}
\begin{tabular}{ccccccccc}
\toprule
\textbf{Dataset} & \textbf{\#Size} & \textbf{\#DB} & \textbf{\#D} & \textbf{\#T/DB} & \textbf{Domain} & \texttt{\textbf{ORDER BY}} & \texttt{\textbf{GROUP BY}} & \texttt{\textbf{NESTED}} \\ \toprule
Spider & 10,181 & 200 & 138 & 5.1 & Cross & 1335 & 1491 & 844 \\
WikiSQL & 80,654 & 26,521 & - & 1 & Cross & 0 & 0 & 0 \\
Advising & 3,898 & 208 & 1 & 10 & Single  & 15 & 9 & 22 \\
BIRD & 12,751 & 95 & 37 & 7.3 & Cross & 2576 & 881 & 0 \\
IMDB & 131 & 1 & 1 & 16 & Single  & 10 & 6 & 1 \\
Yelp & 128 & 1 & 1 & 7 & Single  & 18 & 21 & 0 \\ \midrule
\textbf{\booksqlNormal} & 100k & 1 & 1 & 7 & Single & 17,529 & 11,508 & 4,456 \\ \bottomrule
\end{tabular}
\caption{Comparison of benchmark datasets with  \booksql. \#Size, \#DB, \#D, and \#T/DB represent the numbers of query-SQL pairs, databases, domains, and the averaged number of tables per domain, respectively. The “-” in the \#D column indicates an unknown number of domains. Last 3 columns indicate the query types. Yelp dataset is based on Yelp website, IMDB is based on movie domain and Advising dataset is based on the University Course domain}
\label{tab:statistics of multiple datasets}
\vspace{-4mm}
\end{table*}

\noindent To the best of our knowledge, there is no large-scale dataset in the accounting domain that contains granular records of accounting books used in businesses. To give an idea about the scale of usage of accounting databases: there are around $33$ million small businesses\footnote{\url{https://tinyurl.com/mr3vrptj}} in the US alone. Most of these businesses use accounting software to maintain their books to keep track of their finances, i.e., money-in transactions (e.g., invoice and sales receipt) and money-out transactions (e.g., expense, purchase order, and bill payment).  Additionally, for tax purposes, these books need to follow standard accounting principles like double-entry accounting,\footnote{\url{https://en.wikipedia.org/wiki/Double-entry\_bookkeeping}} hierarchical chart of account structure,\footnote{\url{https://en.wikipedia.org/wiki/Chart\_of\_accounts}} and accrual accounting.\footnote{\url{https://en.wikipedia.org/wiki/Basis\_of\_accounting}} Transactions in the accounting database span across multiple tables. The corresponding SQL queries can involve complex operations such as aggregations, computing distinct counts, and nested queries to extract information from these. For a novice user, this is not an easy task. Moreover, as observed in our initial experiments (Table \ref{tab:sota-models}), existing state-of-the-art (SOTA) \textsql\  models trained on Spider have very poor performance on domain-specific \booksql\  dataset, pointing towards the need for a accounting domain specific dataset which will further lead to the development of SOTA models.  In a nutshell, in this resource paper, we make the following contributions:

\begin{enumerate}
\item We create a new and large-scale \textsql\  financial dataset referred to as \booksql. The dataset consists of a financial-accounts database of 1 million records. The corresponding natural language queries are designed to address various practical intricacies of the accounting domain. \booksqlNormal\  has 100k Query-SQL pairs which is about 1.25 times the existing largest Text-2-SQL dataset: WikiSQL. In particular, for designing the queries, we consulted financial experts to understand various practical use cases. 
\item We run existing state-of-the-art models (including GPT-4) for the \textsql\  task on \booksqlNormal\  to see the performance and analyze the shortcomings of the models trained on existing large-scale datasets such as Spider, pointing towards developing specialized models for this domain. We release the dataset and model code via GitHub: \url{https://github.com/Exploration-Lab/BookSQL}.
\end{enumerate}

\section{Related Work} \label{sec:related}

Due to its importance in practical applications, developing natural language interfaces to databases has been an active area of research. Due to space constraints, we cannot cover all the research, and we refer the reader to the survey by \citet{deng-etal-2022-recent-advances-in-text-to-sql}. We outline some of the main works in this area in this section. Several datasets have been proposed for \textsql\  task in recent years. For example, \textit{Spider} \cite{spider234} dataset has been proposed; it covers 138 different domains. A large-scale dataset, WikiSQL \cite{wikiSQL}, consisting of 24241 Wikipedia tables, has been created. Similarly, Squall \cite{shi-etal-2020-potential}, KaggleDBQA \cite{lee-etal-2021-kaggledbqa}, and BIRD-SQL \cite{li2023llm} datasets have been generated to evaluate the generalization property of models on unseen domains. Domain-specific datasets have also been proposed, such as those based on Yelp and IMDB \cite{imdb}, Advising domain \cite{Advising}, MIMICSQL \cite{MIMICSQL}, SEDE \cite{SEDE}, Restaurants domain \cite{tang:ecml01}, and Academic domain \cite{academic}. The purpose of these datasets is to evaluate the performance of models with a high degree of precision while disregarding the generalization characteristic of the models. 

\noindent\textbf{Comparison.} We compare \booksqlNormal with other popular datasets in Table \ref{tab:statistics of multiple datasets}. As can be observed, \booksqlNormal\  has a much large number of Query-SQL pairs, has a more diverse number of queries in terms of the SQL clauses (e.g., \texttt{ORDER BY}), and involves more complex (and nested) queries. Benchmark dataset such as Spider have a very wide coverage over various domains (138) but very few queries per domain (e.g., average number of queries per domain is 74 in the case of Spider), limiting its performance in a specific domain (see also Table \ref{tab:sota-models}). Moreover, \booksqlNormal\ can be merged with the existing Spider dataset to increase its coverage in the business domain.  

\noindent\textbf{Models.} Various models have been proposed for the \textsql\ task \cite{deng-etal-2022-recent-advances-in-text-to-sql}. Some state-of-the-art models include the non-invasive UniSAr model \cite{unisar} based on Seq2Seq architecture. The model has shown high accuracy on the multi-domain, multi-table Spider dataset. RESDSQL \cite{li2023resdsql} decouples the schema linking and the skeleton parsing for \textsql\ generation. Schema linking identifies the table and columns required for a given question. Skeleton parsing first generates the SQL skeleton and then the final SQL. It achieves SOTA performance on the Spider benchmark. 

\section{\booksqlNormal\ Dataset} \label{sec:corpus}

\begin{table}[t]
\centering
\begin{tabular}{cc}
\toprule
\booksql & \textbf{Stats} \\ 
\midrule
Size of the database &  1 million\\ 
Total Businesses & 27 \\
Size of Question-SQL Pair & 100k \\
Number of Easy SQL & 10,000\\
Number of Medium SQL & 45,000\\
Number of Hard SQL & 45,000\\
\bottomrule
\end{tabular}%
\caption{Statistics of \booksqlNormal\!.}
\label{tab:bookSQL-stats}
\vspace{-5mm}
\end{table}

\begin{figure*}[h]
    \centering
    \includegraphics[scale=0.35]{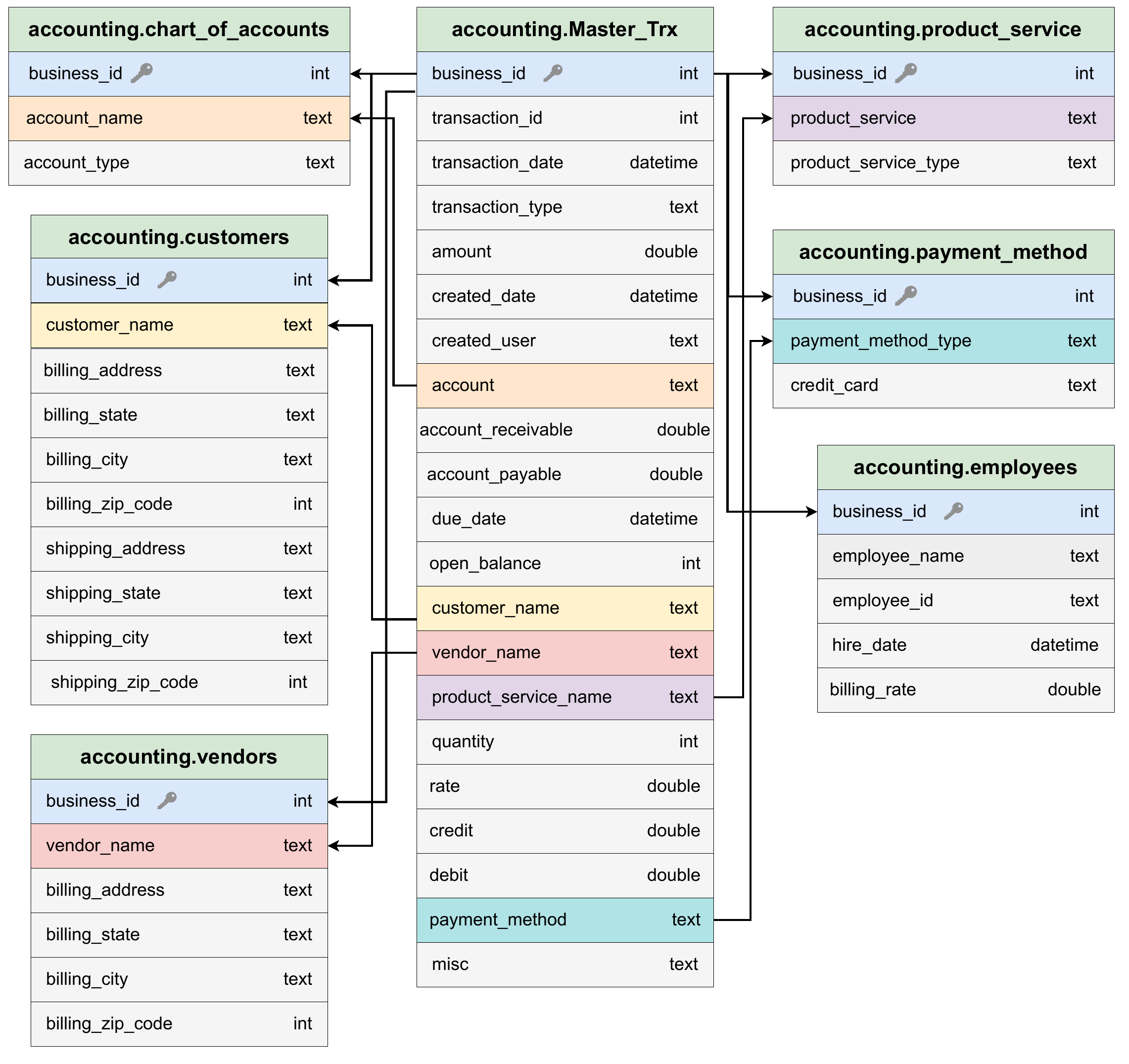}
    \caption{\booksqlNormal\ Database schema} 
    \label{fig:bookSQL_schema}
\end{figure*}

\noindent Given the importance and wide prevalence of business databases across the world, the proposed dataset, \booksqlNormal\ focuses on the finance and accounting domain. 
Accounting databases are used across a wide spectrum of industries like construction, healthcare, retail, educational services, insurance, restaurant, real estate, etc. Business in these industries arranges their financial transactions into their own different set of categories (called a chart of accounts\footnote{https://www.investopedia.com/terms/c/chart-accounts.asp} in accounting terminology). For example, a restaurant business could have categories like advertising, license fees, etc., a real estate brokerage business could have categories like commissions, office supplies, etc. Keeping generalization in mind \booksqlNormal\ dataset includes a variety of businesses from different industries. Hence, a \textsql\ system developed on \booksqlNormal\ will be robust at handling various types of accounting databases. The total size of the dataset is 1 million. The dataset is prepared under financial experts' supervision, and the dataset's statistics are provided in Table \ref{tab:bookSQL-stats}. The dataset consists of 27 businesses, and each business has around 35k - 40k transactions. The distributions of all businesses and their products are shown in Appendix Figure \ref{fig:bookSQL_business_distribution-example} and Figure \ref{fig:bookSQL_business_distribution}.


\subsection{\booksqlNormal\ Tables}
Figure \ref{fig:bookSQL_schema} shows the detailed database schema. The schema is reflective of real-life databases used in the finance and accounting domain. There are seven tables in the \booksqlNormal, namely, Master Transactions, Customer, Employees, Product Service, Vendor, Chart of Account, and Payment Method tables. We arrived at the list of seven tables after examining (and corresponding discussions with finance experts) the databases of several businesses. Given the nature of accounting domain, majority of databases used by businesses across the globe are restricted mainly to these seven tables only. The main table is the ``Master Transaction'' table (e.g., Appendix Table \ref{tab:my-master}), which records money-in transactions (invoice, sales receipt, etc.) and money-out transactions (expense, purchase order, bill payment, etc.) This table also records additional corresponding transaction details, like the customer, vendor, product/service, credit account, debit account, and amount. The ``Chart of accounts'' table (e.g., Appendix Table \ref{tab:my-chart}) contains information on all account names and types. The ``Customer'' table (e.g., Appendix Table \ref{tab:my-customers}) contains all the customer's details, i.e., name, billing, and shipping address. The ``Vendors'' table (e.g., Appendix Table \ref{tab:my-vendor}) contains all the vendor details of all the businesses, i.e., vendor names and billing addresses. The ``Employees'' table (e.g., Appendix Table \ref{tab:my-employee}) contains information about all the business employees. The ``Product service'' table (e.g., Appendix Table \ref{tab:my-product}) contains the details of all the products and services. The ``Payment method'' table (e.g., Appendix Table \ref{tab:my-payment-method}) contains different payment methods the business uses. 


\subsection{Financial Constraints}
For creating the dataset, we took existing accounting databases based on the schema described above and anonymized the names and entries in the tables, i.e., actual names, businesses, and numbers were replaced with fictional ones while adhering to the financial constraints (described next). This is done to maintain the privacy of individuals and businesses. The resulting database is a true reflection of a real-world accounting setting. Accounting databases follow certain accounting rules and financial constraints, which were followed when anonymizing the database. In particular, standard double-entry accounting was followed, which means every entry to an account needs a corresponding and opposite entry to a different account, i.e., debit and credit. So, the sum of debit should always be equal to the sum of credit for every transaction. All seven tables were partitioned by business\_id. For a given transaction\_id, the sum of the credits column should equal the sum of the debits column, and both should equal the amount column in the Master Transactions table. Credit (in the Master Transaction table) should be equal to the product of Quantity and Rate. The chart of accounts was anonymized using the industry-wise list published by a popular CPA.\footnote{\url{https://hectorgarcia.com/resources/}} Business-specific custom fields were anonymized using the examples provided in the help articles of various accounting software. The created database was cross-checked with financial experts to make sure that the created database looked like a real-world accounts database.

\subsection{Dataset Creation and Annotation}
\booksqlNormal\ dataset consists of 100k questions in natural language and their corresponding SQL on multiple tables, covering 27 different businesses. We involved financial experts in the query creation process. We collaborated with two financial experts who have previously been involved in the creation of accounting software. Moreover, these experts have the knowledge and experience in dealing with customer interactions involving account books. The financial experts helped us on a pro bono basis since the creation of \textsql\  system for the accounting domain would help them and their customers.    

\begin{figure}[t]
    \centering
    \includegraphics[scale=0.36]{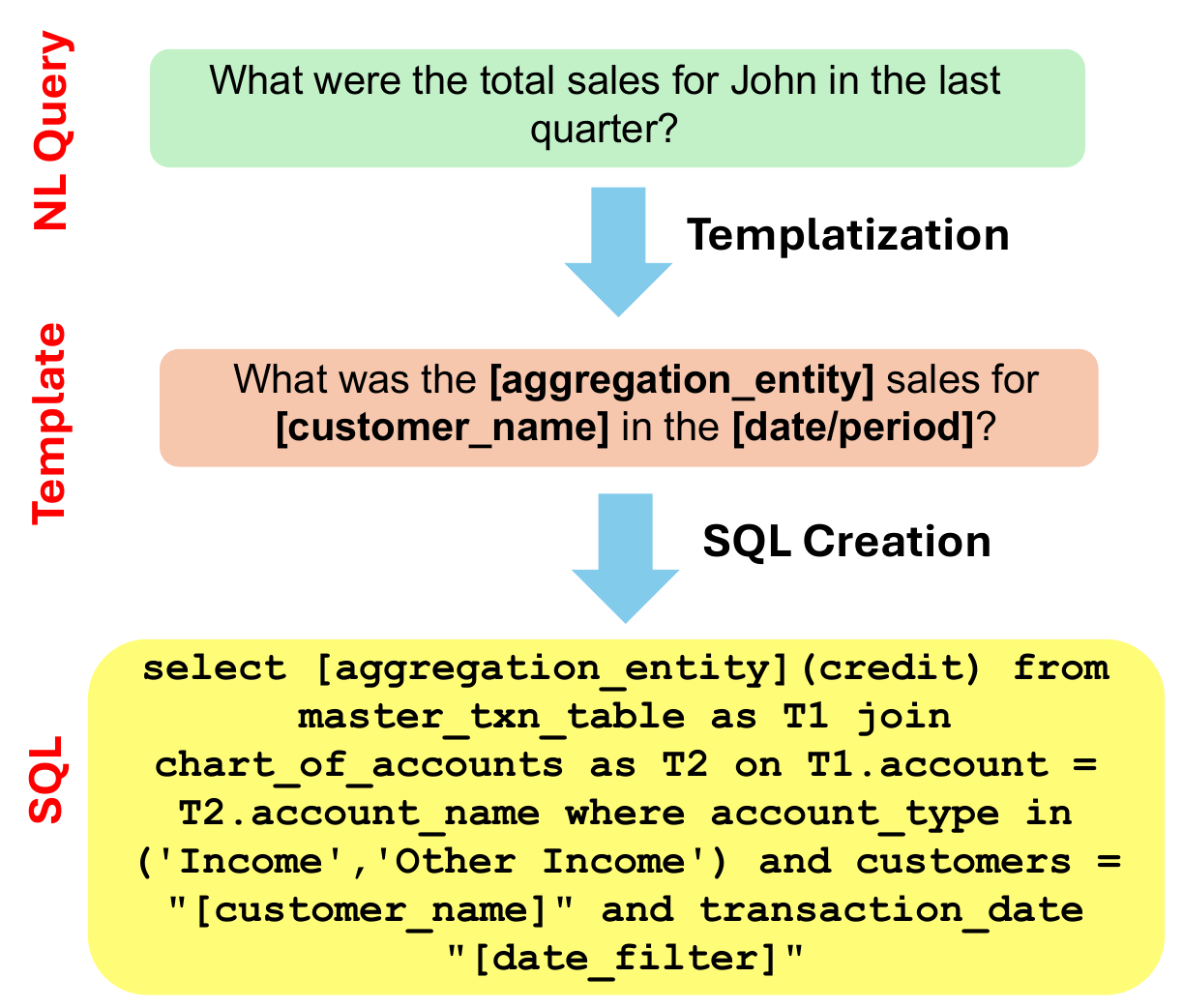}
    \caption{An example showing the pipeline for creating BookSQL dataset. Note, here we can replace \textit{aggregation\_entity} by max, min, total, and average, and \textit{customer\_name} can be replaced with any possible name to get the Question-SQL pair. Similarly, \textit{date/period} can be replaced with \textit{last quarter, this quarter, last month}. }
    \label{fig:pipeline}
\end{figure}

\noindent The question-SQL pair formulation process is as follows. With the help of financial experts, we first created a list of typical questions (based on the account book) that customers (or business people) usually ask or questions about the information that customers are interested in knowing. We tried to keep the questions (queries) as natural as possible to capture real-world scenarios. We relied on the experience of financial experts to keep the list as exhaustive as possible. We also created the corresponding SQL query for each of the natural language queries in the list. The queries in the list were then used to create more queries via the process of templatization. Figure \ref{fig:pipeline} explains the process with help of an example. 



\noindent In order to be as exhaustive as possible, with the help of experts, we arrived at a list of $183$ unique natural language questions that customers typically ask when interacting with accounting databases. These natural language questions were used to create query templates, and this was further used to generate diverse range of Question-SQL pairs in \booksqlNormal. Additionally, we performed a second round of verification of the \booksqlNormal\ corpus and query templates with financial experts to verify the consistency,  veracity, and ensure that the dataset reflected the real-world scenario. Note that existing general \textsql\ datasets (e.g., Spider and WikiSQL) consist of databases from multiple domains and \booksqlNormal\ is focused on the financial domain, hence the number of templates may appear to be less. However, the number of templates is still large when compared across a single domain, for example, to give a rough estimate, Spider dataset uses 5693 templates and spans 138 domains, so a rough estimate of number of templates per domain is about 41 ($\sim 5693/138$). Note that Spider doesn’t provide details about templates for each domain, hence a rough estimate. Moreover, questions in existing \textsql\ datasets (like Spider) are created by students \cite{spider234}, whereas questions in \booksqlNormal\ are created by financial experts who use accounting systems on a regular basis and are well-versed. Although our dataset is small (in terms of total number of templates), it is of high quality and more complex; hence it helps in learning models that would generalize well. Moreover, while experimenting with models, the queries in the test set are based on templates that are not used during training (see section \ref{sec:experiments}). 

\noindent To the best of our knowledge, \booksqlNormal\ is the first \textsql\ dataset to have multi-step questions, which requires nested SQL queries to get the answer. For example - \textit{"What products are selling less than last month/week?"} It would first require computing monthly/weekly product level sales and then comparing each product's current and last month's/week's sales. \booksqlNormal database schema also contains complex column types. Additionally, \booksqlNormal is the first \textsql dataset to have extensive time-based filters like last month, this quarter to date, last financial year, between July to August, this week, yesterday, etc.

\begin{table*}[t]
    \centering
    \renewcommand{\arraystretch}{1}
\setlength\tabcolsep{4pt}
    \begin{tabular}{lp{3cm}p{10cm}}
        \toprule
        \textbf{Complexity} & \textbf{Question} & \textbf{SQL} \\
        \addlinespace[3pt]
        \midrule
        \addlinespace[3pt]
        Easy & What is the balance owned by John?  &  SELECT balance from Customers where customer\_name = 'John'  \\
        \addlinespace[3pt]
        \hline
        \addlinespace[3pt]
        Medium & What is the maximum sales for John in the last month? & SELECT MAX(credit) FROM master\_txn\_table where account\_type in ('Income', 'Other Income') AND customer = 'John' AND month(transaction\_date) = month(current\_timestamp) - 1  \\
        \addlinespace[3pt]
        \hline
        \addlinespace[3pt]
        Hard & What products are selling less than last month? & SELECT A.product\_service, revenue\_this\_month, revenue\_last\_month FROM (SELECT product\_service, SUM(credit) as revenue\_this\_month FROM master\_txn\_table WHERE account\_type in ('Income', 'Other Income') AND month(transaction\_date) = month(current\_timestamp) GROUP BY 1) AS A INNER JOIN (SELECT product\_service, SUM(credit) as revenue\_last\_month FROM master\_txn\_table WHERE account\_type in ('Income', 'Other Income') AND month(transaction\_date) = month(current\_timestamp) - 1 GROUP BY 1) AS B ON A.product\_service = B.product\_service WHERE revenue\_this\_month < revenue\_last\_month \\
        \bottomrule
    \end{tabular}
    \caption{Examples of Question-SQL pairs from \booksqlNormal based on complexity of the query.}
    \label{fig:sql-complexity}
\end{table*}

\subsection{Complexity of SQL in \booksqlNormal}
SQL queries in \booksqlNormal\ are diverse and cover various levels of complexity, i.e., it covers the following operations: SELECT with multiple columns and aggregations, WHERE, GROUP BY, HAVING, ORDER BY, LIMIT, JOIN, INTERSECT, UNION, NOT IN, OR, AND, EXISTS, CONTAINS as well as nested queries. Table \ref{tab:statistics of multiple datasets} shows the comparisons of all \textsql\ datasets. In terms of complexity, \booksqlNormal\ consists of complex SQL queries containing 17,529 ORDER BY, 11,508 GROUP BY, and 4,456 NESTED queries. We further divided all Query-SQL pairs into three categories: Easy, Medium, and Hard, based on the complexity of SQL. Table \ref{fig:sql-complexity} shows examples for each category. Table \ref{tab:bookSQL-stats} shows the main statistics of the \booksqlNormal. \booksqlNormal\ consists of 7,193 \texttt{Hard} SQL queries, making it a more complex, large, and challenging dataset. We used the following criteria to decide on the complexity of a query. 
\begin{itemize}
\item EASY: simple queries with single WHERE condition
\item MEDIUM: multiple conditions in WHERE clause and multiple columns in SELECT clause
\item HARD: Join, Group by, Inner queries, Union, Except as these are hard to predict from Natural Language question.
\end{itemize}

\section{Baseline Models} \label{sec:models}

We benchmark existing state-of-the-art (SOTA) \textsql\ models on \booksqlNormal\ dataset. 



\begin{table*}[t]
    \centering
    \renewcommand{\arraystretch}{1.1}
    \setlength\tabcolsep{6pt}
    \begin{tabular}{p{2.5cm}p{1.75cm}p{1.2cm}p{1.65cm}p{1.5cm}p{1.2cm}cc}
        \toprule
        & \multicolumn{2}{@{}c}{\textbf{Spider}}
        & \multicolumn{5}{c@{}}{\textbf{BookSQL}}\\
        \cmidrule(r){2-3} 
        \cmidrule(r){4-8}
          \multicolumn{1}{@{}c}{\textbf{Model}}   & EMA & EA & EMA & PCM-F1 & EA & BLEU-4 & ROUGE-L   \\ 
        \cmidrule(r){1-1}
        \cmidrule(r){2-3} 
        \cmidrule(r){4-8}
         \multicolumn{1}{@{}c}{SEDE}  & 63.2\% & - & 43.4\% & 0.82 & 44.3\% & 0.69 & 0.83   \\
        \multicolumn{1}{@{}c}{UniSAr} & 70\% & - & 43.0\% & 0.78 & 47.6\% & 0.72 & 0.80 \\
        \multicolumn{1}{@{}c}{RESDSQL} & 80.5\% & 84.1\% & 51.5\% & 0.81 & 54.4\% & 0.74 & 0.81 \\
        \multicolumn{1}{@{}c}{DIN-SQL+GPT4} & 60\% & 85.3\% & 9.3\% & 0.63 & 7.6\% & 0.43 & 0.68  \\
        \multicolumn{1}{@{}c}{DFew+GPT4} & - & - & 47.5\% & 0.89 & 67.2\% & 0.86 & 0.90 \\
\bottomrule
\end{tabular}
\caption{Results on Spider and BookSQL datasets. EMA refers to Exact Match Accuracy, EA refers to Execution Accuracy, and PCM-F1 refers to Partial Component Match F1. DFew+GPT4 refers to Dynamic few-shot prompt+GPT4}
\label{tab:exp-results-booksql-spider}
\end{table*}

\noindent\textbf{SEDE:}
We fine-tuned the SEDE model \cite{hazoom2021text} on the \booksqlNormal\  dataset. SEDE is a T5-based sequence-to-sequence model \cite{raffel2020exploring}. It takes unordered schema items (tables and column names) along with questions as input and generates the corresponding SQL query as output. 

\noindent\textbf{UniSAr:}
We fine-tuned the UniSAr model \cite{unisar} on the BookSQL train dataset, with T5-large as the base language model. UniSAr converts any seq-to-seq language model into a text2sql model by three non-invasive extensions:
(1) Structure Mark to encode database schema in the model input, (2) Constrained Decoding to generate well-structured SQL. For the \booksqlNormal\   dataset, we removed the constrained decoding module of UniSAr, since it did not support the SQL queries with complex grammar present in the \booksqlNormal\   dataset. (3) SQL Completion for completing potential missing JOIN relationships.

\noindent\textbf{RESDSQL:}
RESDSQL \cite{li2023resdsql} decouples the schema linking and the skeleton aware decoding for SQL generation. A cross-encoder is trained to rank the tables and columns required for a given query for schema linking. For SQL generation, a seq-to-seq model with skeleton-aware decoding is used, which first generates an SQL skeleton, and then the model predicts the actual SQL query from it. The masked self-attention method in the decoder allows the first created skeleton to direct the future SQL parsing implicitly. 

\noindent\textbf{DIN-SQL + GPT4:}
We use prompt chaining technique as proposed in \citet{pourreza2023din}. It decomposes \textsql\ task into multiple sub-tasks and then solves each sub-task one by one by prompting GPT4 \cite{achiam2023gpt} with sub-task-specific prompts. It uses the following sub-tasks: 
\begin{enumerate}
\itemsep0em 
    \item \textbf{Schema Linking:} This module identifies references to database tables and columns required to answer the natural language question.
    \item \textbf{Classification and Decomposition:} This module classifies each question into easy, non-nested complex, and nested complex. This signifies the type of SQL query required for the given question.
    \item \textbf{SQL Generation:} This module generates the SQL using the output of previous modules.
    \item \textbf{Self Correction module:} This module is responsible for correcting any minor mistakes in the SQL generated by the previous module.
\end{enumerate}


\noindent Sample prompts for each of these sub-tasks are provided in the Appendix \S\ref{app:din-sql-prompts}. 

\noindent\textbf{Dynamic few-shot prompt + GPT4 (DFew+GPT4):}
We follow a dynamic few-shot prompting technique similar to \citet{sun2023sql}. Firstly, a vector database is created by an embedding train set questions using SentenceTransfomers \textit{all-MiniLM-L6-v2} model.\footnote{https://huggingface.co/sentence-transformers/all-MiniLM-L6-v2} This model is trained on the 1 billion sentence pairs dataset\footnote{https://huggingface.co/blog/1b-sentence-embeddings} and is best suited for generating sentence embeddings. This created embedding database is called trainDB. Then, at inference time, embedding for the test question is created using the same SentenceTransfomers model, and this embedding is used to do ANN (Approximate Nearest Neighbor) search in trainDB to get ten examples from the train set. These ten examples and database schema is used to create the few-shot SQL generation prompt for GPT4. Pseudo-code and sample prompts are provided in Appendix \S\ref{app:dynamic-gpt-prompts}. We use ChromaDB\footnote{https://github.com/chroma-core/chroma} as the underlying vector database and for ANN search.

\section{Experiments, Results and Analysis} \label{sec:experiments}
\subsection{Evaluation Metrics} 
We use the standard evaluation metrics (details in Appendix \ref{app:metrics}) of Exact Match Accuracy (EMA) \cite{spider234}, Execution Accuracy (EA) \cite{spider234}, Partial Component Match F1 (PCM-F1) \cite{SEDE}, BLEU-4 \cite{bleu}, and ROUGE-L \cite{lin-2004-rouge}. 

\subsection{Experimental Setup}
We divide the dataset into $70\%$ train, $10\%$ validation, and $20\%$ test sets based on query templates. The test set contains $14.37\%$ easy, $78.43\%$ medium, and $7.2\%$ hard SQL queries. In order to check the generalization performance, queries in the test set are based on templates that are not used during training. Given limitations on the number of calls to OpenAI GPT4 API, we used a random $10\%$ of \booksqlNormal\ test set for GPT4-based approaches. We provide details about training and hyper-parameters in Appendix \ref{app:hyper-parameters}.

\begin{table}[t]
    \setlength{\tabcolsep}{3.5pt}
    \centering
    \begin{tabular}{p{1cm}cccc}
        \toprule
        \textbf{Query} &  \textbf{SEDE} &  \textbf{UniSAr} &  \textbf{RESDSQL} & \textbf{GPT4}   \\  
        \midrule
         \multicolumn{1}{@{}c}{E} & 100 & 100 & 100 & 100 \\
        \multicolumn{1}{@{}c}{M} & 43.08 & 46.49 & 62.12 & 71.35 \\
        \multicolumn{1}{@{}c}{H} & 15.00 & 12.34 & 15.00 & 22.08 \\
\bottomrule
\end{tabular}
\caption{Execution Accuracy (in \%) of various models on SQL queries of varying complexity. E refers to \textbf{Easy} query, M refers to \textbf{Medium} query and H refers to query with \textbf{Hard} complexity.}
\label{tab:sql-complexity-results}
\end{table}

\begin{table*}[h!]
\centering
\small
\renewcommand{\arraystretch}{0.9}
\setlength\tabcolsep{3pt}
\begin{tabular}{llp{12cm}}
\toprule
1 & {\color[HTML]{3531FF} Question:} & \textbf{What was the average invoice value for Biogenic municipal waste-fueled power generation?} \\
\addlinespace[4pt]
 & {\color[HTML]{3531FF} Gold SQL:} & \textcolor{brown}{\textbf{SELECT}} avg(credit) \textcolor{brown}{\textbf{FROM}} master\_txn\_table \textcolor{brown}{\textbf{WHERE}} transaction\_type = 'invoice' \textcolor{brown}{\textbf{AND}} instr(account, 'Biogenic municipal waste-fueled power generation') \\
 \addlinespace[4pt]
 & {\color[HTML]{3531FF} Few-shot GPT4:} & \textcolor{brown}{\textbf{SELECT}} avg(\textcolor{red}{amount}) \textcolor{brown}{FROM} master\_txn\_table \textcolor{brown}{\textbf{WHERE}} transaction\_type = 'invoice' \textcolor{brown}{\textbf{AND}} \textcolor{red}{product\_service} = 'Biogenic municipal waste-fueled power generation' \xmark \\
 \addlinespace[4pt]
 & {\color[HTML]{3531FF} SEDE:} & \textcolor{brown}{\textbf{SELECT}} avg(\textcolor{green}{credit}) \textcolor{brown}{\textbf{FROM}} master\_txn\_table \textcolor{brown}{\textbf{WHERE}} transaction\_type = 'invoice' \textcolor{brown}{AND} instr(account, 'biogenic municipal waste-fueled power generation') \cmark \\
 \addlinespace[4pt]
 & {\color[HTML]{3531FF} UniSAr:} & \textcolor{brown}{\textbf{SELECT}} avg(credit) \textcolor{brown}{\textbf{FROM}} master\_txn\_table \textcolor{brown}{\textbf{WHERE}} transaction\_type = 'invoice' \textcolor{brown}{\textbf{AND}} \textcolor{red}{product\_service} = 'Biogenic municipal waste-fueled power generation' \xmark \\
  \addlinespace[4pt]
 & {\color[HTML]{3531FF} RESDSQL:} & \textcolor{brown}{\textbf{SELECT}} avg(\textcolor{green}{credit}) \textcolor{brown}{\textbf{FROM}} master\_txn\_table \textcolor{brown}{\textbf{WHERE}} transaction\_type = 'invoice' \textcolor{brown}{\textbf{AND}} instr(account, 'biogenic municipal waste-fueled power generation') \cmark \\
 \addlinespace[3pt]
 \midrule
 \addlinespace[3pt]
2 & {\color[HTML]{3531FF} Question:} & \textbf{What was the last invoice value for Drain cleaning in This week to date?} \\
\addlinespace[4pt]
 & {\color[HTML]{3531FF} Gold SQL:} & \textcolor{brown}{\textbf{SELECT}} max(credit) \textcolor{brown}{\textbf{FROM}} master\_txn\_table \textcolor{brown}{\textbf{WHERE}} transaction\_type = 'invoice' \textcolor{brown}{\textbf{AND}} instr(account,'Drain cleaning') \textcolor{brown}{\textbf{AND}} transaction\_date \textcolor{brown}{\textbf{BETWEEN}} date( current\_date, 'weekday 0', '-7 days') \textcolor{brown}{AND} date( current\_date) \\
 \addlinespace[4pt]
 & {\color[HTML]{3531FF} Few-shot GPT4:} & \textcolor{brown}{\textbf{SELECT}} \textcolor{red}{credit} \textcolor{brown}{\textbf{FROM}} master\_txn\_table \textcolor{brown}{\textbf{WHERE}} transaction\_type = 'invoice' \textcolor{brown}{\textbf{AND}} \textcolor{red}{product\_service} = 'Drain cleaning' \textcolor{brown}{\textbf{AND}} transaction\_date \textcolor{brown}{BETWEEN} date(current\_date, 'weekday 0', '-7 days') \textcolor{brown}{\textbf{AND}} date(current\_date) 
 \textcolor{red}{\textbf{ORDER BY} transaction\_date DESC LIMIT} 1 \xmark  \\
 \addlinespace[4pt]
 & {\color[HTML]{3531FF} SEDE:} & \textcolor{brown}{\textbf{SELECT}} max(credit) \textcolor{brown}{\textbf{FROM}} master\_txn\_table \textcolor{brown}{\textbf{WHERE}} transaction\_type = 'invoice' \textcolor{brown}{\textbf{AND}} \textcolor{red}{customers = 'drain} cleaning' \textcolor{brown}{\textbf{AND}} transaction\_date \textcolor{brown}{\textbf{BETWEEN}} date( current\_date, 'weekday 0', '-7 days') \textcolor{brown}{\textbf{AND}} date( current\_date) \xmark  \\
 \addlinespace[4pt]
 & {\color[HTML]{3531FF} UniSAr:} & \textcolor{brown}{\textbf{SELECT}} max(credit) \textcolor{brown}{\textbf{FROM}} master\_txn\_table \textcolor{brown}{\textbf{WHERE}} transaction\_date \textcolor{brown}{\textbf{BETWEEN}} date ( current\_date , 'weekday 0' , '-7 days' ) \textcolor{brown}{\textbf{AND}} date ( current\_date ) \xmark \\
  \addlinespace[4pt]
 & {\color[HTML]{3531FF} RESDSQL:} & \textcolor{brown}{\textbf{SELECT}} \textcolor{green}{max ( credit )} \textcolor{brown}{\textbf{FROM}} master\_txn\_table \textcolor{brown}{\textbf{WHERE}} transaction\_type = 'invoice' \textcolor{brown}{\textbf{AND}} \textcolor{green}{instr ( account , 'Drain cleaning' )} \textcolor{brown}{\textbf{AND}} transaction\_date \textcolor{brown}{\textbf{BETWEEN}} date ( current\_date , 'weekday 0' , '-7 days' ) \textcolor{brown}{\textbf{AND}} date ( current\_date ) \cmark \\
 \addlinespace[3pt]
 \midrule
 \addlinespace[3pt]
3 & {\color[HTML]{3531FF} Question:} & \textbf{What is my average revenue for Customer Nathan Hernandez in the last 6 years?} \\
\addlinespace[4pt]
 & {\color[HTML]{3531FF} Gold SQL:} & \textcolor{brown}{\textbf{SELECT}} sum(credit)/6 \textcolor{brown}{\textbf{FROM}} master\_txn\_table \textcolor{brown}{\textbf{WHERE}} customers = 'Nathan Hernandez' \textcolor{brown}{\textbf{AND}} strftime('\%Y', transaction\_date) \textgreater{}= strftime('\%Y', current\_date)-6-1 \\
 \addlinespace[4pt]
 & {\color[HTML]{3531FF} Few-shot GPT4:} & \textcolor{brown}{\textbf{SELECT}} sum(credit)/6 \textcolor{brown}{\textbf{FROM}} master\_txn\_table \textcolor{brown}{\textbf{WHERE}} customers = 'Nathan Hernandez' \textcolor{brown}{\textbf{AND}} strftime('\%Y', transaction\_date) \textgreater{}= strftime('\%Y', current\_date)-6-1 \cmark \\
 \addlinespace[4pt]
 & {\color[HTML]{3531FF} SEDE:} & \textcolor{brown}{\textbf{SELECT}} avg(credit) \textcolor{brown}{\textbf{FROM}} master\_txn\_table \textcolor{red}{AS t1 JOIN chart\_of\_accounts AS t2 ON t1.account = t2.account\_name \textcolor{brown}{\textbf{WHERE}} account\_type IN ('income','other income')} \textcolor{brown}{\textbf{AND}} customers = 'nathan hernandez' \textcolor{brown}{\textbf{AND}} transaction\_date \textcolor{brown}{\textbf{BETWEEN}} \textcolor{red}{date(current\_date,'start of year','-6 year') \textcolor{brown}{\textbf{AND}} date(current\_date,'start of year','-1 day')} \xmark \\
  \addlinespace[4pt]
  & {\color[HTML]{3531FF} UniSAr:} & \textcolor{brown}{\textbf{SELECT}} avg(credit) \textcolor{brown}{\textbf{FROM}} master\_txn\_table \textcolor{red}{as T1 \textcolor{brown}{\textbf{JOIN}} chart\_of\_accounts as T2 on T1.account = T2.account\_name \textcolor{brown}{\textbf{WHERE}} account\_type in ( 'Income' , 'Other Income' )} {\textbf{AND}} customers = 'Nathan Hernandez' \textcolor{brown}{\textbf{AND}} transaction\_date \textbf{BETWEEN} \textcolor{red}{date ( current\_date ,'start of year' , '-6 year' ) \textcolor{brown}{\textbf{AND}} date ( current\_date ,'start of year' , '-1 day')} \xmark \\
  \addlinespace[4pt]
 & {\color[HTML]{3531FF} RESDSQL:} & \textcolor{brown}{\textbf{SELECT}} sum (credit) / 6 \textcolor{brown}{FROM} master\_txn\_table \textcolor{brown}{\textbf{WHERE}} customers = 'Nathan Hernandez' \textcolor{brown}{\textbf{AND}} strftime ( '\%Y' , transaction\_date ) \textgreater{}= strftime ( '\%Y' , current\_date ) - 1 \textcolor{red}{\textbf{GROUP BY} strftime ( '\%Y' , transaction\_date )} \xmark \\

\bottomrule
\end{tabular}
\caption{Error analysis for different models on \booksqlNormal test set}
\label{tab:error-analysis}
\end{table*}

\subsection{Results}
Table \ref{tab:exp-results-booksql-spider} and Table \ref{tab:sql-complexity-results} shows the performance of baseline models. Table \ref{tab:exp-results-booksql-spider} shows the performance of SOTA \textsql\ models fine-tuned on the \booksqlNormal\ dataset. RESDSQL performs best as can be observed with regard to exact match accuracy and execution accuracy scores. SEDE and UniSAr have poor exact match and execution accuracy scores. Though \booksqlNormal\ and Spider are not directly comparable, we also include results of the models on the Spider dataset to provide a reference for comparison purposes. As can be observed, the models that perform well on Spider do not have a good performance on \booksqlNormal, indicating the complexity of the dataset. 
Table \ref{tab:exp-results-booksql-spider} shows the in-context learning performance of GPT4 on the BookSQL test set. DIN-SQL+GPT4 could only get $9.3\%$ exact match accuracy, while Dynamic few-shot prompt+GPT4 comes close to the best-fine-tuned model, with exact match accuracy of $47.5\%$ and execution accuracy of $67.2\%$. Table \ref{tab:sql-complexity-results} shows the performance on easy, medium and hard queries. All models have a perfect performance ($100\%$ execution accuracy) on easy queries but struggle with medium and hard queries. 

\subsection{Error Analysis}
We observed that SOTA models fail on queries with date filters, nested queries, distinct aggregations, and domain-specific filters. Table \ref{tab:error-analysis} shows the outputs of models on some examples from the test set. \textbf{DIN-SQL + GPT4} performs very poorly with execution accuracy of 7.6\%. Perhaps, the reason for the bad performance is that it uses the same static chain-of-thought prompt, irrespective of the test question. \booksqlNormal\ questions are very diverse and require domain knowledge. It is impossible to capture this diversity and domain knowledge in only a few examples in the prompt. Due to this, DIN-SQL fails whenever the test question is completely different from the examples provided in the prompt. \textbf{Dynamic few shot prompt + GPT4} model addresses the limitations of DIN-SQL by dynamically selecting a few shot examples for the prompt based on the test question. It significantly improves execution accuracy to 67.2\%. Possibly, the reasons for poor performance are: 1) Getting confused between different columns (like \texttt{WHERE} clause on product\_service vs. account column - see table \ref{tab:error-analysis}), 2) Mixing up credit, debit, and amount columns and using incorrect columns in aggregations, 3) Not able to generate Nested SQL, even when required to answer the test question correctly, 4) failing when domain-specific information is required to generate SQL correctly. For example, transaction\_type filters of the invoice, sales receipt, and purchase order, or account\_type filters of expense, income, account receivable, and account payable are incorrectly applied. 


\noindent\textbf{SEDE} fails to generate correct SQL, possibly due to a lack of question and schema linking in the input to the T5 model. Due to this, it mixes up different columns like customer, vendor, product\_service, and account. \textbf{UniSAr} performs poorly, possibly due to complex queries introduced in \booksqlNormal\ like date filters, nested queries, distinct aggregations, etc. UniSAr introduces constrained grammar-based decoding, which works well for simple queries but fails with such complex queries. \textbf{RESDSQL} is the best-performing model. 
Poor performance is possibly due to: 1) Failure at complex time-based questions like \textit{"What is average revenue for customer X in last 6 years"} (see table \ref{tab:error-analysis}); 2) mixing up of credit and debit columns; 3) failure when distinct aggregations are required like \texttt{COUNT(DISTINCT transaction\_id)}; 4) failure in case of many nested queries. 
\section{Future Directions}
Results show the poor performance of the SOTA models on \booksqlNormal. We outline some of the possible directions for the future to improve performance. 

\noindent\textbf{Multi-task learning:} One could employ a multi-task learning setup, i.e., in addition to optimizing for SQL generation objective, adding other multi-task objectives could help improve the performance on hard SQL queries. These objectives could include (1) nested vs. non-nested SQL classification, (2) distinct keyword classification, and (3) date format classification. 

\noindent\textbf{Pre-training:} For large databases, it is difficult for any model to relate the question tokens with column names when the question might refer to some table cell value. Before the \textsql\ task, one could do pre-training to better understand the question and table relationships. This can be done using mask modeling by defining tasks such as column recovery and column predictions where few tokens could be masked, and the model tries to recover and predict the masked tokens; a similar approach is proposed by \citet{shi2020learning} via the GAP model. 

\noindent\textbf{Multi-step few-shot prompting:} One could also generate SQL in multiple steps using dynamic few-shot prompting instead of generating in a single step.

\noindent\textbf{Value Encoding:} In-context learning models (GPT4) mixes up different columns due to a lack of knowledge about table contents. Adding related table rows in the prompt could alleviate this issue.






\section{Conclusion} \label{sec:conclusion}

In this paper, we propose \booksqlNormal, a \textsql\ dataset that will have broad applications in the finance and accounting domain. The experimental outcomes of several \textsql\ models indicate considerable room for improvement. In the future, we aim to build a more robust model that can handle hard queries and improve performance.


\section*{Limitations}
Since this is a resource paper, we release a large dataset and consequently focus less on modeling the \textsql\ system. We tested existing \textsql\ systems to see how well these fare on the new dataset. The results are indicative of considerable scope for improvement. In the future, we will focus on developing new models with better performance on BookSQL. Moreover, we hope that once the dataset is released, it will foster more research in this domain, resulting in more interesting models. 

\section*{Ethics Statement}
Considering the privacy aspect, we create anonymized entries in the dataset. Moreover, the dataset was verified by financial experts to make sure that the entries adhere to accounting principles and are reflective of real-life scenarios. We will be releasing the dataset publicly for research uses. To the best of our knowledge, we are not aware of any other possible ethical consequences of the proposed dataset.

\bibliography{./sections/references}

\clearpage
\newpage

\appendix
\section*{Appendix} \label{sec:appendix}

\section{Dataset Details}
\begin{itemize}[noitemsep,nosep]
    \item Table \ref{tab:my-master} shows master transaction table. 
    \item Table \ref{tab:my-chart} shows Chart of account table. 
\item Table \ref{tab:my-customers} shows Customer table 
\item Table \ref{tab:my-vendor} shows Vendor table. 
\item Table \ref{tab:my-employee} shows Employee table. 
\item Table \ref{tab:my-product} shows Product Service table. 
\item Table \ref{tab:my-payment-method} shows Payment Methods table. 
\end{itemize}

\begin{figure*}[h!]
    \centering
    \includegraphics[scale=0.60]{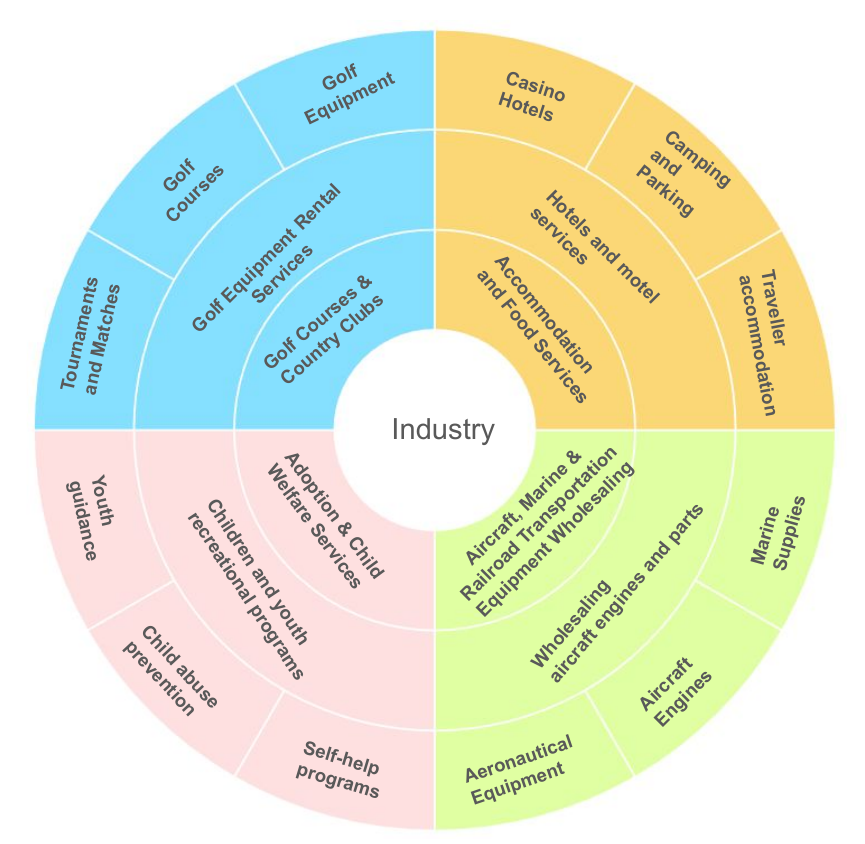}
    \caption{Sample \booksqlNormal\ Business Distribution. The middle section shows the sample set of businesses, inner section shows the industries associated with the corresponding business and outer most section shows the corresponding product of the business. This chart is made with the information available at: \url{https://www.ibisworld.com/united-states/list-of-industries/}.} 
    \label{fig:bookSQL_business_distribution-example}
\end{figure*}

\begin{figure*}[h!]
    \centering
    \includegraphics[scale=0.40]{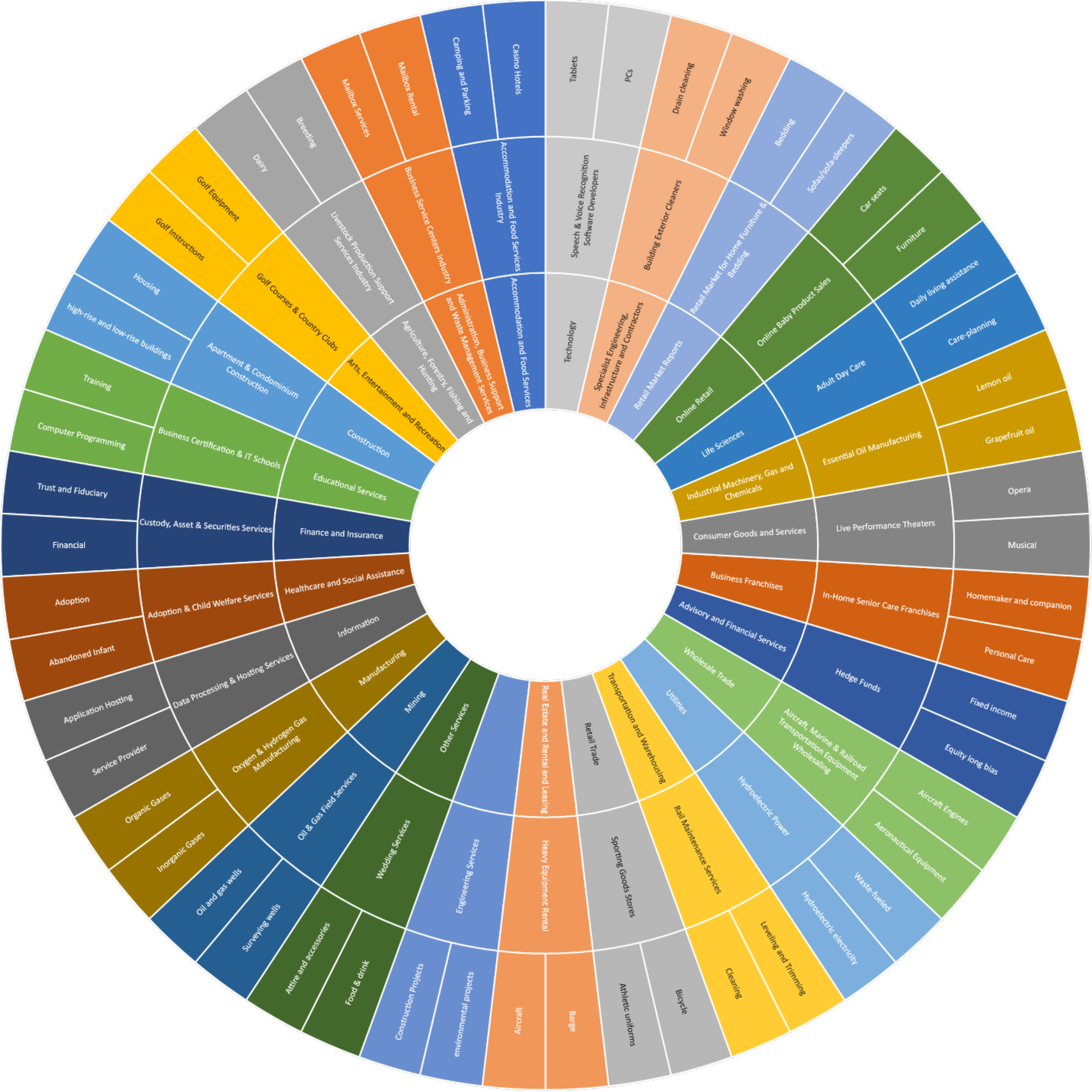}
    \caption{\booksqlNormal Business Distribution. Here, inner circle indicates the industries , middle circle shows the sets of businesses associated to respective industry , and the outer most circle indicate corresponding product of the business. This chart is made with the information available at: \url{https://www.ibisworld.com/united-states/list-of-industries/}.}
    \label{fig:bookSQL_business_distribution}
\end{figure*}

\paragraph{Distribution of Businesses in \booksqlNormal}

The distributions of all businesses and their products are shown in Figure \ref{fig:bookSQL_business_distribution}. Each industry is represented in the inner circle layer, which is followed by its businesses (in the middle circle) and its products in the outer circle. Each industry comprises multiple businesses, and each business consists of multiple products and services. 

\onecolumn

\section{Dynamic Few-shot Prompt + GPT4} \label{app:dynamic-gpt-prompts}
Pseudo-code for dynamic few-shot train example selection for a given test question:

\begin{footnotesize}
\begin{lstlisting}[language=Python, frame=single]
from langchain.embeddings.huggingface import HuggingFaceEmbeddings

from langchain.prompts.example_selector import MaxMarginalRelevanceExampleSelector

from langchain.vectorstores import Chroma

example_selector = 
    MaxMarginalRelevanceExampleSelector.from_examples(
        examples, 
        HuggingFaceEmbeddings(model_name="all-MiniLM-L6-v2"), 
        Chroma, 
        k=10,
        input_keys=["input"]
    )
\end{lstlisting}
\end{footnotesize}

\subsection{Example Prompt}
\textit{
Database schema:
\newline
Table master\_txn\_table, columns = [*, Transaction\_ID, Transaction\_DATE, Transaction\_TYPE, Amount, CreatedDATE, CreatedUSER, Account, AR\_paid, AP\_paid, Due\_DATE, Open\_balance,                             Customers, Vendor, Product\_Service, Quantity, Rate, Credit, Debit, payment\_method, Misc]
\newline
Table chart\_of\_accounts, columns = [*, Account\_name, Account\_type]
Table customers, columns = [*, customer\_name, customer\_full\_name, Billing\_address, Billing\_city, Billing\_state, Billing\_ZIP\_code, Shipping\_address, Shipping\_city, Shipping\_state, Shipping\_ZIP\_code, Balance]
\newline
Table employees, columns = [*, Employee\_name, Employee\_ID, Hire\_date, Billing\_rate, Deleted]
\newline
Table products, columns = [*, Product\_Service, Product\_Service\_type]
\newline
Table vendors, columns = [*, Vendor\_name, Billing\_address, Billing\_city, Billing\_state, Billing\_ZIP\_code, Balance]
\newline
Table payment\_method, columns = [*, Payment\_method, Credit\_card]
\newline
\newline
Foreign\_keys = [master\_txn\_table.Account = chart\_of\_accounts.Account\_name, master\_txn\_table.Customers = customers.customer\_name, master\_txn\_table.Vendor = vendors.Vendor\_name, master\_txn\_table.Product\_Service = products.Product\_Service, master\_txn\_table.payment\_method = payment\_method.payment\_method]
\newline 
\newline
Following are the example of questions and corresponding SQL queries. 
\newline
<<10 Few shot examples from train set>>
\newline
\newline
Translate following question to SQL query. 
\newline
Input: How much open credit does customer Ronald Bailey have? 
\newline
Output: SELECT 
}

\section{DIN-SQL+GPT4 Prompts} \label{app:din-sql-prompts}
Following section shows the sample prompts used in different DIN-SQL modules. For brevity, we have added only 1 few shot example in these sample prompts. Though in practice, 5-10 few shot examples are used and is mentioned at the end of prompt in \textbf{<< >>}.

\subsection{Schema Linking Prompt}
Table master\_txn\_table, columns = [*, Transaction\_ID, Transaction\_DATE, Transaction\_TYPE, Amount, CreatedDATE, CreatedUSER, Account, AR\_paid, AP\_paid, Due\_DATE, Open\_balance, \
                            Customers, Vendor, Product\_Service, Quantity, Rate, Credit, Debit, payment\_method, Misc]
\newline
Table chart\_of\_accounts, columns = [*, Account\_name, Account\_type]
\newline
Table customers, columns = [*, customer\_name, customer\_full\_name, Billing\_address, Billing\_city, Billing\_state, Billing\_ZIP\_code, Shipping\_address, Shipping\_city, Shipping\_state, Shipping\_ZIP\_code, Balance]
\newline
Table employees, columns = [*, Employee\_name, Employee\_ID, Hire\_date, Billing\_rate, Deleted]
\newline
Table products, columns = [*, Product\_Service, Product\_Service\_type]
\newline
Table vendors, columns = [*, Vendor\_name, Billing\_address, Billing\_city, Billing\_state, Billing\_ZIP\_code, Balance]
\newline
Table payment\_method, columns = [*, Payment\_method, Credit\_card]
\newline
Foreign\_keys = [master\_txn\_table.Account = chart\_of\_accounts.Account\_name, master\_txn\_table.Customers = customers.customer\_name, master\_txn\_table.Vendor = vendors.Vendor\_name, master\_txn\_table.Product\_Service = products.Product\_Service, master\_txn\_table.payment\_method = payment\_method.payment\_method]
\newline
\newline
Q: How much open credit does customer Ronald Bailey have?
\newline
S: select sum(open\_balance) from ( select distinct transaction\_id, open\_balance from master\_txn\_table where customers = 'Ronald Bailey')
\newline
A: Let’s think step by step. In the question "How much open credit does customer Ronald Bailey?", we are asked:
    "How much open credit", so we need column = [master\_txn\_table.open\_balance]
    "open credit does customer Ronald Bailey", so we need column = [master\_txn\_table.transaction\_id, master\_txn\_table.customers]
    Based on the columns and tables, we need these Foreign\_keys = [].
    Based on the tables, columns, and Foreign\_keys, The set of possible cell values are = [Ronald Bailey]. So the Schema\_links are:
    Schema\_links: [master\_txn\_table.open\_balance, 
 master\_txn\_table.customers, master\_txn\_table.transaction\_id, Ronald Bailey]

\textsc{<<9 more few-shot examples>>}

\subsection{Classification prompt}
Q: What are my transactions MTD? 
\newline
schema\_links: [master\_txn\_table.transaction\_id, master\_txn\_table.amount, master\_txn\_table.transaction\_date]
\newline
A: Let’s think step by step. The SQL query for the question "What are my transactions MTD?" needs these tables = [master\_txn\_table], so we don't need JOIN.
Plus, it doesn't require nested queries with (INTERSECT, UNION, EXCEPT, IN, NOT IN), and we need the answer to the questions = [""].
So, we don't need JOIN and don't need nested queries, then the the SQL query can be classified as "EASY".
\newline
Label: "EASY"
\newline
\newline
Q: How many products are never sold with total value higher than 5? 
\newline
schema\_links: [Product\_Service.transaction\_id, master\_txn\_table.transaction\_type]
\newline
A: Let’s think step by step. The SQL query for the question "How many products are never sold with total value higher than 5?" needs these tables = [Product\_Service,master\_txn\_table], so we need JOIN.
Plus, it requires nested queries with (INTERSECT, UNION, EXCEPT, IN, NOT IN) or inner query inside from clause, and we need the answer to the questions = ["products that are sold with total value higher than 5"].
So, we need JOIN and need nested queries, then the the SQL query can be classified as "NESTED".
\newline
Label: "NESTED"
\newline
\newline
Q: YTD, what was our smallest expense?
\newline
schema\_links = [master\_txn\_table.account = chart\_of\_accounts.account\_name, master\_txn\_table.credit, master\_txn\_table.transaction\_date, master\_txn\_table.account\_type, master\_txn\_table.debit]
\newline
A: Let’s think step by step. The SQL query for the question "YTD, what was our smallest expense?" needs these tables = [master\_txn\_table,chart\_of\_accounts], so we need JOIN.
Plus, it doesn't need nested queries with (INTERSECT, UNION, EXCEPT, IN, NOT IN), and we need the answer to the questions = [""].
So, we need JOIN and don't need nested queries, then the the SQL query can be classified as "NON-NESTED".
\newline
Label: "NON-NESTED"

\textsc{<<7 more few-shot examples>>}

\subsection{SQL Generation}
\subsubsection{Easy Prompt}
Q: "How much open credit does customer Ronald Bailey?"
\newline
Schema\_links: [master\_txn\_table.open\_balance, master\_txn\_table.transaction\_id, master\_txn\_table.customers,Ronald Bailey]
\newline
SQL: select sum(open\_balance) from ( select distinct transaction\_id, open\_balance from master\_txn\_table where customers = 'Ronald Bailey')

\textsc{<<4 more few-shot examples>>}

\subsubsection{Non-Nested Complex Prompt}
Q: "How many Traveller accomodation did we sell to Ethan Walker today?"
\newline
Schema\_links: [master\_txn\_table.quantity ,master\_txn\_table.customers, master\_txn\_table.product\_service, master\_txn\_table.transaction\_type, master\_txn\_table.transaction\_date]
\newline
A: Let’s think step by step. For creating the SQL for the given question, we need to join these tables = []. First, create an intermediate representation, then use it to construct the SQL query.
Intermediate\_representation: select sum(master\_txn\_table.quantity) from master\_txn\_table where master\_txn\_table.customers = 'Ethan Walker' and master\_txn\_table.product\_service = 'Traveller accomodation' and master\_txn\_table.trasaction\_type in ('invoice','sales receipt') and master\_txn\_table.transaction\_date BETWEEN date(current\_date) AND date(current\_date)
\newline
SQL: select sum(quantity) from master\_txn\_table where customers = \"Ethan Walker\" and product\_service = \"Traveller accomodation\" and trasaction\_type in ('invoice','sales receipt') and transaction\_date BETWEEN date(current\_date) AND date(current\_date)

\textsc{<<9 more few-shot examples>>}

\subsubsection{Nested Complex Prompt}
Q: "How many products are never sold with total value higher than 5?"
Schema\_links: [master\_txn\_table.product\_service, master\_txn\_table.transaction\_type, master\_txn\_table.credit, product\_service.*]
\newline
A: Let's think step by step. "How many products are never sold with total value higher than 5?" can be solved by knowing the answer to the following sub-question "Show me all the products which are never sold with total credit value higher than 5?".
The SQL query for the sub-question "Show me all the products which are never sold with total credit value higher than 5?" is SELECT count(*) FROM Product\_Service WHERE product\_service NOT IN ( SELECT product\_service FROM master\_txn\_table WHERE transaction\_type in ('invoice','sales receipt') group by product\_service  having sum(credit)>5)
So, the answer to the question "How many products are never sold with total value higher than 5?" is =
Intermediate\_representation: SELECT count(Product\_Service.*) FROM Product\_Service WHERE Product\_Service.product\_service NOT IN ( SELECT master\_txn\_table.product\_service FROM master\_txn\_table WHERE master\_txn\_table.transaction\_type in ('invoice','sales receipt') group by master\_txn\_table.product\_service  having sum(master\_txn\_table.credit)  >  5)
\newline
SQL: SELECT count(*) FROM Product\_Service WHERE product\_service NOT IN ( SELECT product\_service FROM master\_txn\_table WHERE transaction\_type in ('invoice','sales receipt') group by product\_service  having sum(credit)  >  5)

\textsc{<<9 more few-shot examples>>}

\subsection{Self Correction Prompt}
For the given question, use the provided tables, columns, foreign keys, and primary keys to fix the given SQLite SQL QUERY for any issues. If there are any problems, fix them. If there are no issues, return the SQLite SQL QUERY as is.
\newline
Use the following instructions for fixing the SQL QUERY:
\newline
1) Use the database values that are explicitly mentioned in the question.
\newline
2) Pay attention to the columns that are used for the JOIN by using the Foreign\_keys.
\newline
3) Use DESC and DISTINCT when needed.
\newline
4) Pay attention to the columns that are used for the GROUP BY statement.
\newline
5) Pay attention to the columns that are used for the SELECT statement.
\newline
6) Only change the GROUP BY clause when necessary (Avoid redundant columns in GROUP BY).
\newline
7) Use GROUP BY on one column only.

\section{Evaluation Metrics} \label{app:metrics}
The following standard metrics are used:
\begin{itemize}
\itemsep0em 
\item \textbf{Exact Match Accuracy \cite{spider234}:} Both predicted and the Gold SQL are decomposed into different SQL components like SELECT, WHERE, GROUP BY, etc. Predicted SQL is marked as correct if all SQL components exactly match with the Gold SQL.
\item \textbf{Execution Accuracy \cite{spider234}:} Output of predicted SQL is the same as Gold SQL's output on execution against the database.
\item \textbf{Partial Component Match F1 \cite{SEDE}:} Both the predicted query and the gold query are parsed into tress using JSqlParser\footnote{https://github.com/JSQLParser/JSqlParser}. These two parsed trees are compared, and an aggregated score is calculated based on the number of matching sub-trees.
\item \textbf{BLEU-4 \cite{bleu}:} It measures the number of matching n-grams between the predicted and the Gold SQL.
\item \textbf{ROUGE-L \cite{lin-2004-rouge}:} It is based on the longest common sub-sequence (LCS) between the predicted and the Gold SQL. A longer shared sequence indicates more similarity between the predicted and the Gold SQL.
\end{itemize}

\section{Training Details and Hyper-parameters} \label{app:hyper-parameters}
All experiments were done on a single NVIDIA A10G Tensor Core GPU.

For SEDE, we used T5-Large as the base seq-to-seq model, with a learning rate of $5e-5$ with 15 epochs and batch size of 6. For decoding, a beam size of 6 was used, with max decoding steps of 250.

For UniSAr, we use T5-Large as a base language model with a learning rate of 1e-5 and max tokens is 1024. We adopt the polynomial\_decay with 5,000 warmup updates. The dropout rate is 0.1. Optimizer is Adam with the default parameters. The max-update is set to 10,000. Empirically, the model obtained the best performance about 10 $\sim$ 15 epochs in \booksqlNormal. The Fairseq dynamically tunes the batch size to realize higher GPU utilization.

For RESDSQL, we used settings recommended by the original paper and code. The Schema Item Classifier module used a RoBERTa-large model with a learning rate of $1e-5$ and an effective batch size of 32 (using gradient accumulation). topk\_table\_num value of 4 and topk\_column\_num value of 8 were used. For the text2sql module, a T5-large model was used with a learning rate of $5e-5$ and an effective batch size of 32 (using gradient accumulation). Beam search decoding was used with num\_beams set to 8 and num\_return\_sequences set to 8.

For DIN-SQL+GPT4 and Dynamic few shot prompt + GPT4, we used OpenAI GPT4 API with following settings: 
\textit{n = 1, temperature=0.0, max\_tokens=600, top\_p = 1.0, frequency\_penalty=0.0, presence\_penalty=0.0}. Given limitations on the number of calls to OpenAI GPT4 API, we used a random 10\% of BookSQL test set for GPT4-based approaches.


\begin{table}[h]
    \tiny
    \centering
    \renewcommand{\arraystretch}{1.5}
    \setlength\tabcolsep{5pt}
    \begin{tabular}{p{0.5cm}p{0.5cm}p{0.5cm}p{0.5cm}p{0.5cm}p{0.5cm}p{0.5cm}p{0.5cm}p{0.5cm}p{0.5cm}p{0.5cm}p{0.5cm}p{0.5cm}p{0.5cm}p{0.5cm}p{0.5cm}p{0.5cm}p{0.5cm}p{0.5cm}}
        \toprule
    \textbf{busin-ess Id} & \textbf{Trans-action ID} & \textbf{Trans-action date} & \textbf{Trans-action type} & \textbf{Amount} & \textbf{Creat-ed date} & \textbf{Creat-ed user} & \textbf{Acco-unt} & \textbf{A/R paid} & \textbf{A/P paid} & \textbf{Due date} & \textbf{Open balance} & \textbf{Cust-omer name} & \textbf{Vendor name} & \textbf{Prod-uct Service} & \textbf{Quan-tity} & \textbf{Rate} & \textbf{Credit} & \textbf{Debit} \\
    \midrule
    4 & 1867 & 2022-08-31 & credit card credit & 999.58 & 2023-05-11 & Joshua Hudson & Visa & - & - & 2023-09-17 & 232.85 & - & - & - & - & - & - & 999.58 \\
    4 & 1867 & 2022-08-31 & credit card credit & 999.58 & 2023-05-11 & Joshua Hudson & Savings & - & - & 2023-09-17 & 232.85 & - & - & - & - & - & 999.58 & - \\
    4 & 1716 & 2022-08-15 & Bill & 784.19 & 2023-05-14 & Joshua Hudson & Accounts Payable (A/P) & - & unpaid & 2023-07-12 & 539.03 & - & Jade Barnett & - & - & - & 784.19 & - \\
    4 & 1716 & 2022-08-15 & Bill & 784.19 & 2023-05-14 & Joshua Hudson & Lawyer & - & unpaid & 2023-07-12 & 539.03 & - & Jade Barnett & - & - & - & - & 784.19 \\
    4 & 1818 & 2022-08-17 & Payment & 2204 & 2022-11-22 & Joshua Hudson & prepaid expenses & paid & - & 2023-06-25 & 1841.82 & Andrew Rose & - & - & - & - & - & 2204 \\
    4 & 1818 & 2022-08-17 & Payment & 2204 & 2022-11-22 & Joshua Hudson & Accounts Receivable (A/R) & paid & - & 2023-06-25 & 1841.82 & Andrew Rose & - & - & - & - & 2204 & - \\
\bottomrule
\end{tabular}
\caption{Master Transaction Table}
\label{tab:my-master}
\end{table}

\begin{table}[h]
    \footnotesize
    \centering
    \begin{tabular}{llll}
        \toprule
\textbf{Business Id} & \textbf{Account name} & \textbf{Account Full Name} & \textbf{Account type} \\ 
\midrule
2 & Accumulated Depreciation & Accumulated Depreciation & Fixed Asset \\
2 & Furniture and Equipment & Furniture and Equipment & Fixed Asset \\ 
2 & Payroll Liabilities & Payroll Liabilities & Other Current Liability \\ 
2 & Opening Balance Equity & Opening Balance Equity & Equity \\ 
2 & Owners Draw & Owners Draw & Equity \\ 
2 & Owners Equity & Owners Equity & Equity \\ 
2 & Accounting Service Income & Accounting Service Income & Income \\ 
2 & Consulting Income & Consulting Income & Income \\ 
2 & Tax Preparation Services Income & Tax Preparation Services Income & Income \\ 
\bottomrule
\end{tabular}%
\caption{Chart of Account}
\label{tab:my-chart}
\end{table}

\begin{table}[h]
    \tiny
    \centering
    \renewcommand{\arraystretch}{1.5}
    \setlength\tabcolsep{5.5pt}
    \begin{tabular}{p{0.8cm}p{1cm}p{1cm}p{1.5cm}p{1cm}p{0.7cm}p{0.7cm}p{1.5cm}p{1cm}p{0.7cm}p{0.7cm}p{0.7cm}}
        \toprule
\textbf{Business Id} & \textbf{Customer name} & \textbf{Customer full name} & \textbf{Billing address} & \textbf{Billing city} & \textbf{Billing state} & \textbf{Billing ZIP code} & \textbf{Shipping address} & \textbf{Shipping city} & \textbf{Shipping state} & \textbf{Shipping ZIP code} & \textbf{Balance} \\ \midrule
2 & Valerie Kline & Valerie Kline & 5120 Shelia Valleys Suite 824 & New Cynthiaburgh & AL & 18662 & 40109 Pamela Extension & West Patrickville & TN & 21599 & 67.32 \\ 
2 & Greg Cardenas & Greg Cardenas & 59614 Margaret Roads & Transide & OK & 72668 & 3758 Savage Garden Suite 126 & Seanbury & VT & 7317 & 167.00 \\ 
2 & Mr. Zachary Levy & Mr. Zachary Levy & 30939 Brandon Ford Suite 571 & South Joanntown & NV & 71097 & 2752 Austin Brooks Suite 864 & Stoutville & MI & 51839 & 69.34 \\
2 & Taylor Hughes & Taylor Hughes & 7945 Soto Point & Monica-mouth & ND & 46573 & 04225 Edwards Valley Suite 176 & Taylorton & DE & 516 & 169.34 \\ 
2 & Jodi Bishop & Jodi Bishop & 850 Brent Parks & Shieldsberg & AL & 7549 & 1558 Brown Hills & South Robert & ID & 94105 & 799.37 \\ 
2 & Andrew Flores & Andrew Flores & 3141 Jamie Isle Apt. 494 & South Nicholasmouth & OH & 71180 & 4662 Peters Parkways Suite 775 & Desireebury & AR & 34741 & 79.37 \\
2 & Earl Lee & Earl Lee & 017 Lisa Skyway & Lake Kristineburgh & AL & 79642 & 8285 Thornton Motorway Suite 926 & Lawsonville & ID & 12448 & 84.75 \\ 
2 & Thomas Jackson & Thomas Jackson & 09653 Christian Stravenue & North Johntown & MS & 43479 & 1205 Shawna Fork Suite 756 & Tracymouth & TX & 77146 & 684.7 \\ 
2 & Jason Johnson & Jason Johnson & 46474 Alan Cove Suite 685 & Michaelside & VT & 50177 & Unit 0702 Box 5832 & DPO & AA & 15548 & 747.44 \\ 
2 & Craig Greer & Craig Greer & 496 Moreno Brooks & Lake Katrinamouth & NM & 14367 & USNV Gutierrez & FPO & AP & 71930 & 47.34 \\ 
2 & Jeffrey Fisher & Jeffrey Fisher & 632 Robert Plains Apt. 260 & Woodardville & LA & 86396 & 46671 Joseph Flat Apt. 818 & Sweeneyshire & DC & 70691 & 5829.13 \\ 
\bottomrule
\end{tabular}%
\caption{Customer Table}
\label{tab:my-customers}
\end{table}

\begin{table}[h]
    \footnotesize
    \centering
    \begin{tabular}{llp{2.5cm}llll}
        \toprule
\textbf{Business Id} & \textbf{Vendor name} & \textbf{Billing address} & \textbf{Billing city} & \textbf{Billing state} & \textbf{Billing ZIP code} & \textbf{Balance} \\ 
\midrule
2 & Shelly Ramos & 82768 Dawn Crescent & West Cynthia & WY & 39877 & 4042.15 \\
2 & Jade Barnett & 782 Mitchell Camp Suite 676 & Grahambury & KS & 80370 & 12949.89 \\ 
2 & Nicole Jordan & 14959 Mccullough Green Suite 029 & East Kevinfurt & WI & 42930 & 5294.89 \\ 
2 & Adam Pena & 192 Brenda Gardens & Erinmouth & IA & 93008 & 6949.89 \\
2 & Jeffrey Roman & 784 Cameron Parks Apt. 902 & North Gloriafurt & AR & 48141 & 7299.89 \\
2 & Zachary Butler & 61717 Christopher Cliffs Apt. 122 & Port Joshua & MT & 44164 & 465.09 \\ 
2 & Taylor Moses & 19368 Jenny Courts Apt. 094 & Kerristad & OR & 25430 & 65.09 \\ 
2 & John Russo & Unit 6387 Box 0856 & DPO & AA & 73133 & 1538.8 \\ 
2 & Robert Phillips & USCGC Steele & FPO & AA & 91533 & 55388.8 \\ 
\bottomrule
\end{tabular}%
\caption{Vendor Table}
\label{tab:my-vendor}
\end{table}

\begin{table}[h]
    \footnotesize
    \centering
    \begin{tabular}{llllll}
        \toprule
\textbf{Business Id} & \textbf{Employee name} & \textbf{Employee ID} & \textbf{Hire date} & \textbf{Billing rate} & \textbf{Deleted} \\ 
\midrule
2 & Stephanie Baker & STE123 & 07/17/2022 & -- & No \\ 
2 & Julia Rivera & JUL456 & 07/31/2002 & -- & No \\ 
2 & Valerie Kline & VAL232 & 04/15/2012 & -- & Yes \\ 
2 & Greg Cardenas & GRE443 & 08/27/2013 & -- & No \\ 
2 & Mr. Zachary Levy & ZAC998 & 01/28/2000 & -- & Yes \\ 
2 & Taylor Hughes & TAY009 & 07/17/2022 & -- & Yes \\ 
2 & Jodi Bishop & JOD778 & 12/27/2016 & -- & Yes \\ 
2 & Andrew Flores & AND667 & 05/20/2018 & -- & No \\ 
2 & Earl Lee & EAR221 & 08/19/2002 & -- & No \\ 
\bottomrule
\end{tabular}%
\caption{Employee Tables}
\label{tab:my-employee}
\end{table}

\begin{table}[h]
    \footnotesize
    \centering
    \begin{tabular}{lll}
        \toprule
\textbf{Business Id} & \textbf{Product\_service} & \textbf{Product\_Service\_type} \\ 
\midrule
2 & Hours & Service \\ 
2 & Services & Service \\ 
2 & Design & Service \\ 
2 & Installation & Service \\ 
2 & Lighting & Service \\ 
2 & Maintenance \& Repair & Service \\ 
2 & Refunds \& Allowances & Service \\
\bottomrule
\end{tabular}%
\caption{Product Service Table}
\label{tab:my-product}
\end{table}

\begin{table}[h]
    \footnotesize
    \centering
    \begin{tabular}{lll}
        \toprule
\textbf{Business Id} & \textbf{Payment method} & \textbf{Credit card} \\
\midrule
1 & Cash & No \\ 
1 & Check & No \\ 
1 & Visa & Yes \\ 
1 & MasterCard & Yes \\ 
1 & Discover & Yes \\ 
1 & American Express & Yes \\ 
1 & Diners Club & Yes \\
\bottomrule
\end{tabular}%
\caption{Payment Methods}
\label{tab:my-payment-method}
\end{table}

\end{document}